\documentclass[letterpaper, 10 pt, conference]{ieeeconf}  

\IEEEoverridecommandlockouts                              

\usepackage{graphics} 
\usepackage{epsfig} 
\usepackage{mathptmx} 
\usepackage{times} 
\usepackage{amsmath} 
\usepackage{amssymb}  
\usepackage{booktabs}

\title{\LARGE \bf
Training-Free Object-Agnostic Jam Detection in Fulfillment Centers
}

\author{Ruiliang Liu, Tina Dongxu Li, Joshua Migdal, Fernando Ruch, Kenneth Meszaros, Moses Trevor Dardik \\
Amazon, USA\\
\{ruilianl, dxl, jmigdal, ruchf, mesza, tdardik\}@amazon.com}

\begin{document}

\maketitle

\begin{abstract}
In fulfillment centers, diverse objects move continuously from inbound to outbound operations and can become jammed due to excessive conveyor friction, incorrect orientation, or mechanical failures. Traditional jam detection approaches rely on object detection models to identify objects, followed by tracking algorithms (such as IoU overlap and Kalman filtering) to monitor motion over time. This pipeline requires thousands of manual annotations, consuming approximately two weeks of effort, and is limited to annotated object classes.

We present a training-free, object-agnostic jam detection method that eliminates the need for labeled data. Our approach uniformly samples reference points within the monitoring region when no objects are present. As objects occlude these points, we detect motion. When a sufficient fraction remains occluded beyond a temporal threshold, we classify the event as a jam. Unlike conventional point tracking---which treats occlusion as a failure case---our approach repurposes occlusion as a detection signal, monitoring whether reference points remain persistently occluded rather than tracking where they move.

Our experimental evaluation on 1,069 videos demonstrates that AllTracker achieves 100.00\% precision and 93.33\% F1 score, significantly outperforming classical sparse tracking methods while maintaining training-free deployment. This approach offers three key advantages: (1) no training data or manual annotations, (2) object-agnostic generalization to arbitrary object types, and (3) significantly reduced development time.
\end{abstract}


\section{Introduction}

Jam detection is a critical operational challenge in fulfillment centers, where totes and packages become obstructed on conveyors due to excessive friction, incorrect orientation, or mechanical failures.

\subsection{Traditional Tracking-by-Detection Approaches}

Current jam detection methods rely on object detection models followed by tracking algorithms. Prior work uses RTMDet, YOLO, and RT-DETR models trained on thousands of manually annotated images to detect totes, boxes, envelopes, and trays, then associates detections across frames using Intersection over Union (IoU) overlap.

These tracking-by-detection methods suffer from two fundamental limitations: (1) they require extensive manual annotation, consuming significant time and resources, and (2) they are constrained to trained object classes---novel object types (such as customer-provided containers) cannot be detected.

Our key insight is a novel repurposing of point tracking. Conventional point trackers follow the \emph{motion} of visible feature points---occlusion is treated as a nuisance. We invert this paradigm: rather than tracking \emph{where} points go, we monitor \emph{whether} uniformly sampled reference points remain occluded over time. This shift from motion estimation to visibility reasoning enables training-free, object-agnostic jam detection without semantic understanding.

\subsection{Proposed Point-Based Approach}

We present a training-free, object-agnostic alternative leveraging point tracking for jam detection. Before objects enter the area of interest (AOI), we uniformly sample reference points across the monitoring region. When a sufficient fraction remains occluded beyond temporal threshold $X$ seconds, we classify the event as a jam.

Our experimental evaluation on 1,069 videos demonstrates that AllTracker achieves 100.00\% precision and 93.33\% F1 score, providing robust performance without requiring training data or manual annotations.

\section{Related work}

\subsection{Point tracking}

Point tracking estimates trajectories of salient points across image sequences. Early work on optical flow [1,2] led to the Lucas--Kanade (LK) method, which became foundational for sparse tracking. The Kanade--Lucas--Tomasi (KLT) tracker [3,4] extended this with corner-based feature selection, enabling robust tracking under partial occlusion.

Subsequent methods improved handling of larger motions and long-term consistency [5-7]. Recent deep learning approaches like CoTracker3 [8] provide multi-point consistency under occlusion, while dense trackers like AllTracker [9] track all pixels simultaneously, offering robustness in texture-poor regions. Our approach leverages both sparse (KLT, CoTracker3) and dense (AllTracker) trackers to monitor reference point visibility for jam detection.

\subsection{Multi-object tracking (MOT)}

Multi-object tracking estimates object trajectories while maintaining consistent identities. Traditional MOT follows tracking-by-detection, using models like YOLO, RTMDet, or RT-DETR with data association methods including bipartite matching, Hungarian algorithm, and Kalman filtering [10-12]. Recent deep learning methods integrate learned appearance representations and end-to-end architectures [13-15], advancing performance on benchmarks like MOT Challenge and KITTI.

However, MOT-based jam detection requires thousands of annotated images and is limited to trained object classes. Our point-based approach eliminates object-level detection, providing a training-free, object-agnostic alternative.

\subsection{Jam detection in fulfillment and transportation systems}

Jam detection has been studied in traffic monitoring and industrial conveyor systems. In fulfillment centers, existing systems use RTMDet, YOLO, and RT-DETR models with IoU-based tracking [16]. While effective for trained classes, these require substantial annotation and cannot generalize to novel objects. Our work formulates jam detection as a point visibility problem, enabling training-free deployment and object-agnostic generalization. Section IV demonstrates this approach achieves 100.00\% precision and 93.33\% F1 score on 1,069 videos while eliminating annotation requirements.

\section{Approach}

\subsection{Problem Formulation}

Given a fixed camera observing an area of interest (AOI), our goal is to detect sustained obstruction events (``jams'') without explicit object detection or object-level tracking. We formulate the problem in terms of point visibility over time, where persistent occlusion of reference points indicates prolonged blockage of the AOI.

\subsection{Point Initialization in an Area of Interest (AOI)}

At initialization, we uniformly sample a set of reference points $\{p_i\}_{i=1}^{N}$ within the AOI. These points serve as virtual probes of scene visibility. Point initialization is performed when the AOI is unoccupied or assumed to represent the static background.

\subsection{Point Tracking}

Each reference point is tracked across frames to monitor its visibility. In the sparse tracking setting, we employ Kanade--Lucas--Tomasi (KLT) [3,4] for classical gradient-based tracking and CoTracker3  for long-term multi-point consistency under occlusion. In the dense tracking setting, we use AllTracker  to track all points simultaneously, providing a full-frame motion field. Tracking provides both the updated location of each point and a confidence or visibility measure used to determine occlusion.

\subsubsection{Why Sparse vs Dense}

Sparse tracking offers computational efficiency and interpretability, while dense tracking provides improved robustness in texture-poor regions. Our framework is compatible with both, differing only in how point motion and visibility are estimated.

\subsection{Occlusion and Motion Reasoning}

After tracking, each reference point's visibility is assessed over time using the outputs provided by each tracker. CoTracker and AllTracker directly output visibility information for each tracked point, which is used to determine occlusion status. A point is considered occluded when the tracker indicates it is not visible.

For KLT, we adapt the occlusion detection logic to better suit its texture-based tracking characteristics. Since KLT aims to track points on textured surfaces, a jam is identified by zero-velocity persistence of tracked features coupled with low photometric error. This combination indicates that the object is visible but stationary, which is the signature of a jam condition rather than simple occlusion.

For jam detection, we define a temporal threshold $T = 5$ seconds and a spatial threshold $\tau$. A jam is declared when a sufficient fraction of points remain occluded for longer than 5 seconds:
\begin{equation}
\frac{1}{N} \sum_{i=1}^{N} \mathbb{1}(p_i \text{ occluded for } \geq T) > \tau
\end{equation}
where $N$ is the total number of tracked points, and $\mathbb{1}(\cdot)$ is the indicator function.

Our framework is tracker-agnostic, allowing either sparse or dense tracking depending on computational constraints and scene characteristics. By combining point visibility over time with temporal aggregation, the method detects both short-term motion and sustained blockage without requiring object-level detection or classification.

\section{Experiments}

\subsection{Datasets}

We selected 1,069 videos with frame counts ranging from 17 to 2,583 to evaluate model performance across varying video lengths. To assess robustness under diverse operational conditions, our dataset includes videos from static cameras, cameras mounted on moving conveyors, and cameras subjected to vibration. The dataset is proprietary and not publicly available. All videos were collected from historical fulfillment center operations and manually labeled for evaluation purposes. The distribution of categories and jam occurrences is presented in Table~\ref{tab:dataset}.

The positive ratio of 0.75\% reflects the composition of the curated evaluation set and does not represent the empirical frequency of jam occurrences in production fulfillment center operations.

\begin{figure}[htbp]
  \centering
  \begin{minipage}{0.32\columnwidth}
    \centering
    \includegraphics[width=\linewidth]{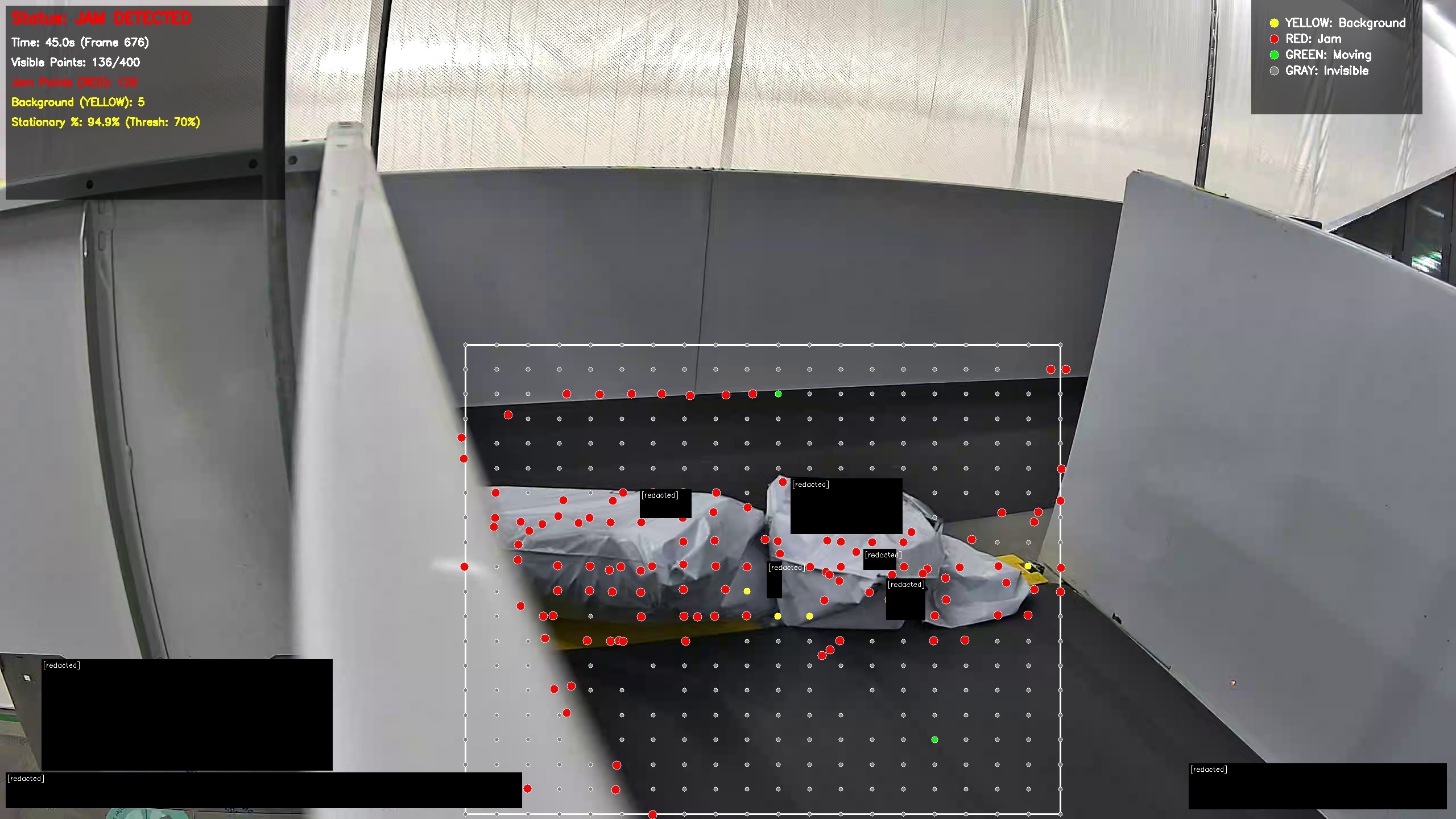}
    \caption{KLT package true positive}
    \label{fig:klt_tp}
  \end{minipage}
  \hfill
  \begin{minipage}{0.32\columnwidth}
    \centering
    \includegraphics[width=\linewidth]{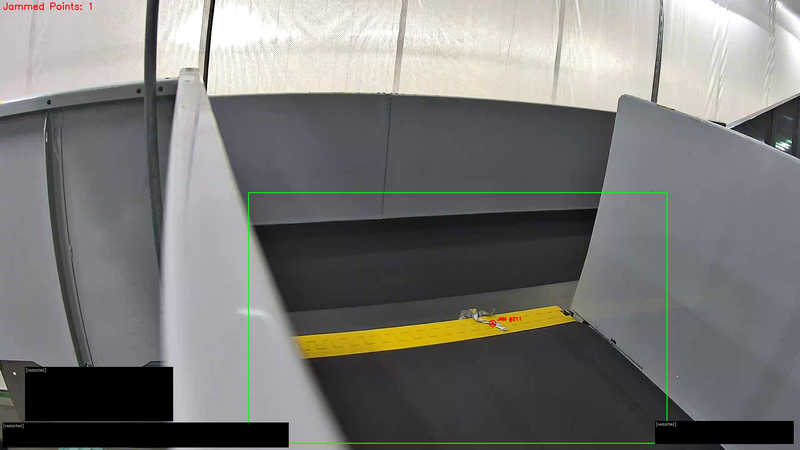}
    \caption{CoTracker3 debris true positive}
    \label{fig:cotracker3_tp}
  \end{minipage}
  \hfill
  \begin{minipage}{0.32\columnwidth}
    \centering
    \includegraphics[width=\linewidth]{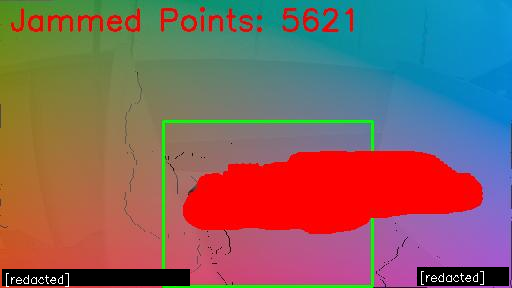}
    \caption{AllTracker package true positive}
    \label{fig:alltracker_tp}
  \end{minipage}
\end{figure}

\begin{table}[htbp]
\caption{Category-wise distribution}
\label{tab:dataset}
\centering
\begin{tabular}{lcccc}
\toprule
\textbf{Category} & \textbf{Total} & \textbf{Jam} & \textbf{Non-Jam} & \textbf{Positive} \\
 & \textbf{Count} & \textbf{Count} & \textbf{Count} & \textbf{Ratio (\%)} \\
\midrule
Debris & 1 & 1 & 0 & 100.00 \\
Tote & 2 & 1 & 1 & 50.00 \\
Package & 1,066 & 6 & 1,060 & 0.56 \\
\bottomrule
\end{tabular}
\end{table}

\subsection{Experimental Setup}

All experiments were conducted on NVIDIA L4 GPUs with 24GB memory in a cloud environment using pre-recorded video data. Each video was resized such that the longer dimension was set to 512 pixels, with the shorter dimension rescaled proportionally to maintain the original aspect ratio. For sparse point trackers (KLT and CoTracker3), reference points were initialized using a uniform $20\times20$ grid sampled within the region of interest (ROI), resulting in 400 tracked points per video.

\subsection{Models}

We evaluated three distinct point tracking approaches representing different paradigms:

\textbf{KLT (Kanade-Lucas-Tomasi)} -- A classical sparse point tracker based on optical flow. KLT detects ``good features to track'' using the Shi-Tomasi corner detector, then tracks them frame-to-frame using the Lucas-Kanade optical flow algorithm.

\textbf{CoTracker3} -- A modern deep learning-based sparse point tracker developed by Meta AI. It employs a transformer-based architecture to track multiple points jointly across video frames, modeling correlations between points to improve robustness.

\textbf{AllTracker} -- A deep learning-based dense point tracking framework designed for universal point tracking across arbitrary video frames. It generalizes across diverse scenes and motion types by combining global context modeling with local motion reasoning.

\subsection{Jam Detection Logic}

We define Regions of Interest (ROI) on the conveyor system and monitor point occlusion patterns within these regions. A jam event is triggered when points in the ROI remain occluded for more than 5 seconds. If the occluded points become visible again later in the video, the jam event is cleared. Only occlusions lasting more than $T = 5$ seconds and persisting until the end of the video are classified as confirmed jam events.

This criterion reflects a practical labeling convention: since jams in our dataset are not manually annotated frame-by-frame, we rely on the video structure itself as a proxy for ground truth. A video ending with sustained occlusion provides strong evidence that the jam was unresolved, whereas transient occlusions that clear before the video ends are treated as non-jam events. This approach avoids the need for precise temporal annotations while remaining consistent with the operational definition of a jam as a sustained, unresolved blockage.

\subsection{Model Performance Comparison}

Table~\ref{tab:detection_performance} and Table~\ref{tab:speed_performance} present the performance comparison of the three point tracking methods for jam detection. All performance metrics are reported for the jam class.

\begin{table}[!t]
\caption{Detection performance comparison}
\label{tab:detection_performance}
\centering
\begin{tabular}{lcccc}
\toprule
\textbf{Model} & \textbf{Accuracy} & \textbf{Precision} & \textbf{Recall} & \textbf{F1 Score} \\
 & \textbf{(\%)} & \textbf{(\%)} & \textbf{(\%)} & \textbf{(\%)} \\
\midrule
KLT & 88.03 & 2.38 & 37.50 & 4.48 \\
CoTracker3 & 86.34 & 3.38 & 62.50 & 6.41 \\
\textbf{AllTracker} & \textbf{99.91} & \textbf{100.00} & \textbf{87.50} & \textbf{93.33} \\
\bottomrule
\end{tabular}
\end{table}

\begin{table}[!t]
\caption{Inference speed comparison}
\label{tab:speed_performance}
\centering
\begin{tabular}{lcc}
\toprule
\textbf{Model} & \textbf{Point Tracking} & \textbf{Jam Detection} \\
 & \textbf{Speed (FPS)} & \textbf{Speed (FPS)} \\
\midrule
\textbf{KLT} & \textbf{654.86} & \textbf{2,320.92} \\
CoTracker3 & 3.27 & 1,459.45 \\
AllTracker & 26.00 & 274.09 \\
\bottomrule
\end{tabular}
\end{table}

\begin{figure}[htbp]
  \centering
  \begin{minipage}{0.32\columnwidth}
    \centering
    \includegraphics[width=\linewidth]{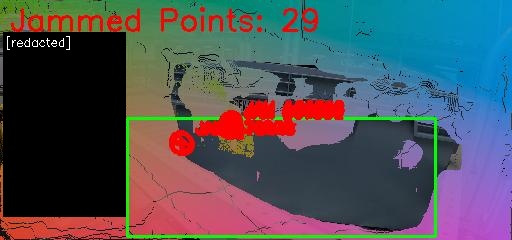}
    \caption{AllTracker tote true positive}
    \label{fig:alltracker_fp}
  \end{minipage}
  \hfill
  \begin{minipage}{0.32\columnwidth}
    \centering
    \includegraphics[width=\linewidth]{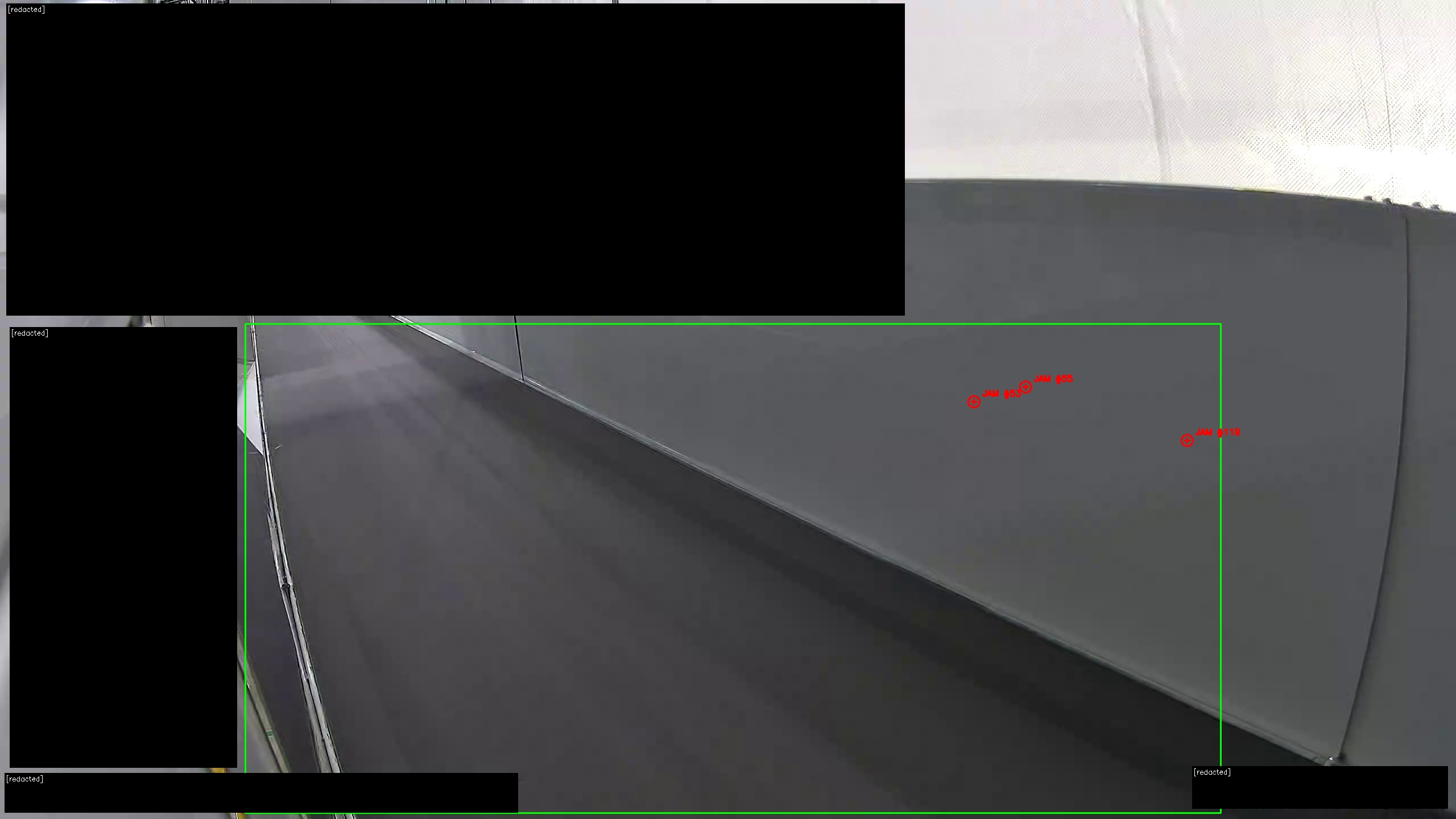}
    \caption{CoTracker3 false positive}
    \label{fig:cotracker3_fp}
  \end{minipage}
  \hfill
  \begin{minipage}{0.32\columnwidth}
    \centering
    \includegraphics[width=\linewidth]{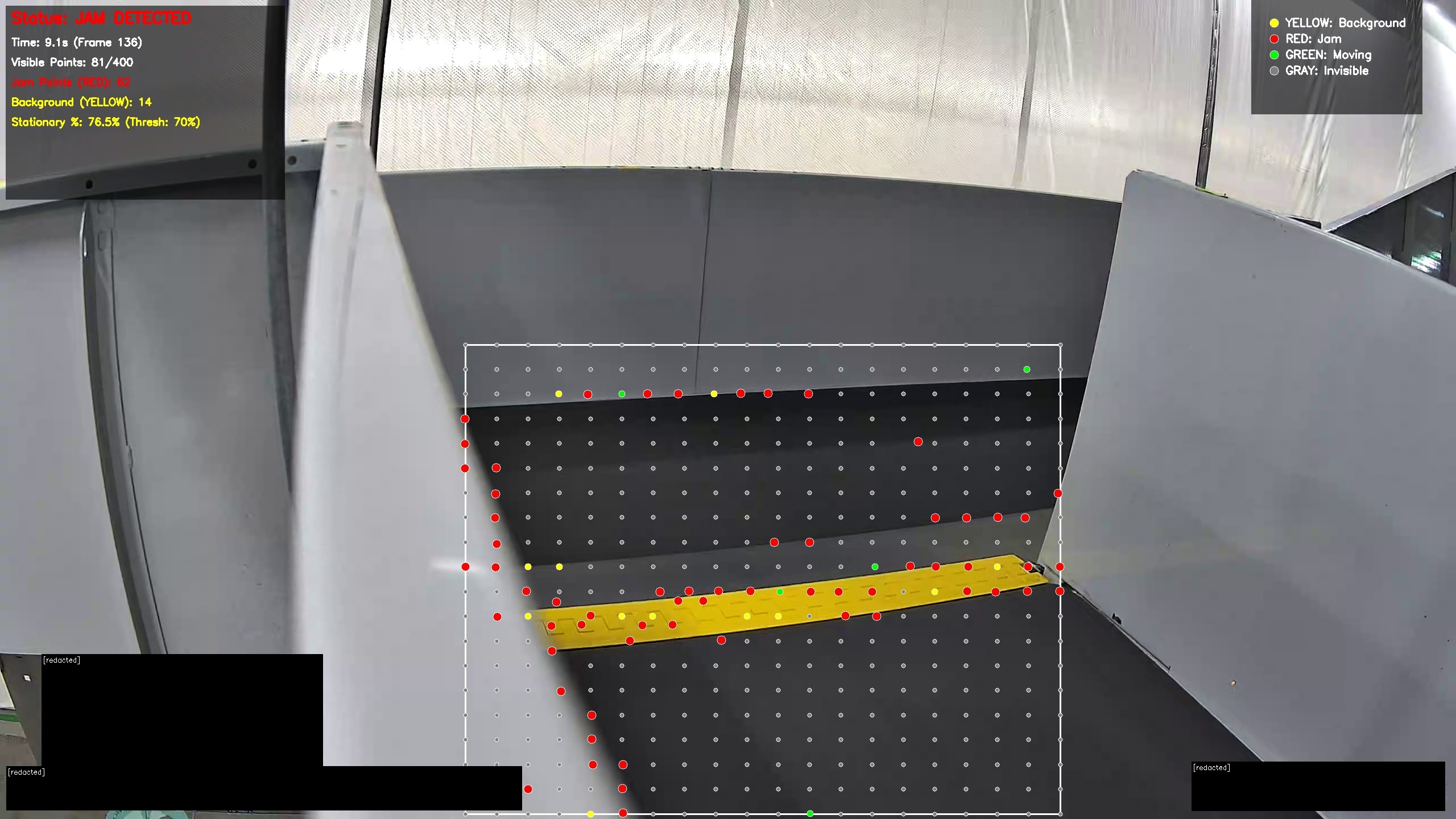}
    \caption{KLT false positive}
    \label{fig:klt_fp}
  \end{minipage}
\end{figure}

Figures~\ref{fig:klt_tp} through~\ref{fig:klt_fp} illustrate representative examples of true positive and false positive detections for each tracking method. Since AllTracker is a dense point tracker that visualizes all tracked pixels, the dense point overlay may obscure the underlying obstruction. To address this, Figures~\ref{fig:klt_tp} and~\ref{fig:alltracker_tp} show the same jam event detected by both KLT and AllTracker, allowing readers to identify the actual obstruction in the sparse KLT visualization before examining the dense AllTracker representation.

AllTracker demonstrates superior detection performance across all metrics, achieving 100.00\% precision, 87.50\% recall, and 93.33\% F1 score. This substantially outperforms both classical sparse tracking (KLT: 4.48\% F1) and modern sparse deep learning approaches (CoTracker3: 6.41\% F1), validating the effectiveness of dense point tracking for training-free jam detection.

\subsection{Analysis}

The superior performance of AllTracker can be attributed to fundamental differences in training methodology. Both KLT and CoTracker3 were designed to track corner-like features and consequently struggle with flat regions, simple edges, and noisy backgrounds. In contrast, AllTracker was trained on diverse point distributions, enabling robust tracking across all point types. This explains its superior precision (100.00\%) and F1 score (93.33\%).

The only jam event AllTracker failed to detect was a debris-type obstruction where the object was exceptionally small. When resized from 512 to 1024 pixels, AllTracker successfully captured the debris jam, demonstrating sensitivity to spatial resolution. AllTracker achieves 26.00 FPS, significantly faster than CoTracker3 (3.27 FPS) while maintaining substantially higher accuracy. Although KLT operates at 654.86 FPS, its poor F1 score (4.48\%) makes it unsuitable for production deployment.

\section{Limitations}

While our point-based tracking approach offers a training-free, object-agnostic solution for jam detection, several limitations warrant discussion:

\textbf{Dependence on initial AOI state:} The method assumes the AOI is unoccupied during point initialization. Objects present during initialization may lead to incorrect baseline establishment.

\textbf{Sensitivity to camera motion:} The approach requires a fixed camera with stable viewpoint. Camera motion, vibration, or perspective changes would cause reference points to appear displaced, potentially triggering false jam detections.

\textbf{Threshold parameter tuning:} The method relies on temporal threshold $T$ (occlusion duration) and spatial threshold $\tau$ (fraction of occluded points). These parameters must be tuned for specific deployment scenarios and may not generalize across different AOI sizes, object densities, or operational contexts.

\textbf{Limited semantic understanding:} By design, the approach operates without object-level detection or classification. While this provides generality and computational efficiency, it sacrifices semantic understanding of jam causes.

Our evaluation is limited by the small number of positive jam events ($n = 8$), which prevents robust statistical analysis. Although AllTracker achieved 100.00\% precision and 87.50\% recall, the confidence intervals are wide given the sample size---missing one additional jam would reduce recall to 75.00\%. Larger-scale validation with 50--100+ jam events is needed to establish reliable confidence intervals and enable statistically rigorous comparisons between tracking methods.

\section{Conclusion}

We presented a training-free, object-agnostic approach to jam detection that monitors reference point visibility, eliminating the need for object detection pipelines requiring thousands of annotated images.

\subsection{Key Contributions}

Our approach offers four primary contributions:

\textbf{1. Training-free deployment:} By eliminating labeled training data, our method avoids the annotation process that typically consumes two weeks and thousands of images, enabling deployment without data collection or model training.

\textbf{2. Object-agnostic generalization:} Unlike detection-based approaches constrained to specific object classes, our point-based formulation detects jams caused by any object type by reasoning about scene visibility rather than object identity.

\textbf{3. Tracker flexibility:} Our framework supports both sparse tracking (KLT, CoTracker3) for computational efficiency and dense tracking (AllTracker) for robustness in texture-poor regions.

\textbf{4. Novel repurposing of point tracking:} Conventional point trackers follow the \emph{motion} of visible points---occlusion is treated as a failure case. We invert this paradigm: rather than tracking \emph{where} points go, we monitor \emph{whether} reference points remain occluded over time, enabling training-free, object-agnostic jam detection without semantic understanding.

\subsection{Practical Impact}

Our retrospective evaluation on 1,069 historical videos demonstrates that AllTracker achieves 100.00\% precision and 93.33\% F1 score, significantly outperforming classical and modern sparse tracking methods. If deployed, fulfillment centers could rapidly deploy this high-accuracy system across diverse locations without waiting for data collection and model training cycles.

\section*{Acknowledgments}

The authors acknowledge the use of a generative AI language tool for drafting and light editing assistance in portions of this manuscript, including the Abstract, Introduction, Related Work, Limitations, and Contributions. All AI-assisted content was reviewed, verified, and approved by the authors. The authors bear full responsibility for the accuracy, originality, and integrity of all content presented in this manuscript.

\section*{References}

{
\small

[1] B. K. P. Horn and B. G. Schunck, ``Determining optical flow,'' {\it Artificial Intelligence}, vol. 17, no. 1-3, pp. 185--203, 1981.

[2] J. L. Barron, D. J. Fleet, and S. S. Beauchemin, ``Performance of optical flow techniques,'' {\it International Journal of Computer Vision}, vol. 12, no. 1, pp. 43--77, 1994.

[3] B. D. Lucas and T. Kanade, ``An iterative image registration technique with an application to stereo vision,'' in {\it Proc. IJCAI}, pp. 674--679, 1981.

[4] J. Shi and C. Tomasi, ``Good features to track,'' in {\it Proc. IEEE CVPR}, pp. 593--600, 1994.

[5] S. Baker and I. Matthews, ``Lucas--Kanade 20 years on: A unifying framework,'' {\it Int. J. Comput. Vis.}, vol. 56, no. 3, pp. 221--255, 2004.

[6] D. Sun, X. Yang, M.-Y. Liu, and J. Kautz, ``PWC-Net: CNNs for optical flow using pyramid, warping, and cost volume,'' in {\it Proc. IEEE CVPR}, pp. 8934--8943, 2018.

[7] T. Teed and J. Deng, ``RAFT: Recurrent all-pairs field transforms for optical flow,'' in {\it Proc. ECCV}, pp. 402--419, 2020.

[8] N. Karaev et al., ``CoTracker3: Simpler and better point tracking by pseudo labelling real videos,'' arXiv:2410.11831, 2024.

[9] A. W. Harley et al., ``AllTracker: Efficient dense point tracking at high resolution,'' in {\it Proc. IEEE ICCV}, 2025.

[10] D. Reid, ``An algorithm for tracking multiple targets,'' {\it IEEE Trans. Autom. Control}, vol. 24, no. 6, pp. 843--854, 1979.

[11] L. Zhang, Y. Li, and R. Nevatia, ``Global data association for multi-object tracking using network flows,'' in {\it Proc. IEEE CVPR}, pp. 1--8, 2008.

[12] A. Bewley et al., ``Simple online and realtime tracking,'' in {\it Proc. IEEE ICIP}, pp. 3464--3468, 2016.

[13]   N. Wojke, A. Bewley, and D. Paulus, ``Simple online and realtime tracking with a deep association metric,'' in {\it Proc. IEEE ICIP}, pp. 3645--3649, 2017.

[14] A. Milan et al., ``MOT16: A benchmark for multi-object tracking,'' arXiv:1603.00831, 2016.

[15] P. Sun et al., ``Scalability in perception for autonomous driving: Waymo Open Dataset,'' in {\it Proc. IEEE CVPR}, pp. 2446--2454, 2020.

[16] F. Halawa, S. Sadeghzadeh, and R. Mohammed, ``Real-time computer vision system for monitoring conveying systems,'' {\it Int. J. Adv. Manuf. Technol.}, vol. 137, pp. 1--15, 2025.

}

\end{document}